\documentclass{ecai} 

\usepackage{latexsym}
\usepackage{amssymb}
\usepackage{amsmath}
\usepackage{amsthm}
\usepackage{booktabs}
\usepackage{enumitem}
\usepackage{graphicx}
\usepackage{color}

\usepackage{array}
\usepackage{multirow}
\usepackage{soul}
\usepackage{caption}
\usepackage{xcolor}
\newcommand{\BibTeX}{B\kern-.05em{\sc i\kern-.025em b}\kern-.08em\TeX}
\newcommand{\CellWithForceBreak}[2][c]{\begin{tabular}[#1]{@{}c@{}}#2\end{tabular}}

\begin{document}

%%%%%%%%%%%%%%%%%%%%%%%%%%%%%%%%%%%%%%%%%%%%%%%%%%%%%%%%%%%%%%%%%%%%%%%%

\begin{frontmatter}

%%% Use this command to specify your submission number.
%%% In doubleblind mode, it will be printed on the first page.

\paperid{432} 

%%% Use this command to specify the title of your paper.

\title{Interpretation of the Intent Detection Problem as Dynamics in a Low-dimensional Space}

\author[A]{\fnms{Eduardo}~\snm{ Sanchez-Karhunen}\orcid{0000-0003-0136-3332}\thanks{Corresponding Author. Email: fesanchez@us.es.}}
\author[A]{\fnms{Jose F.}~\snm{Quesada-Moreno}\orcid{0000-0001-7458-5855}}
\author[A]{\fnms{Miguel A.}~\snm{Guti\'errez-Naranjo}\orcid{0000-0002-3624-6139}} 

\address[A]{University of Seville, Department of Computer Science and Artificial Intelligence. Seville, Spain}

%%% Use this environment to include an abstract of your paper.

\begin{abstract} 
\noindent Intent detection is a text classification task whose aim is to recognize and label the semantics behind a users' query. It plays a critical role in various business applications. The output of the intent detection module strongly conditions the behavior of the whole system. This sequence analysis task is mainly tackled using deep learning techniques. Despite the widespread use of these techniques, the internal mechanisms used by networks to solve the problem are poorly understood. Recent lines of work have analyzed the computational mechanisms learned by RNNs from a dynamical systems perspective.
In this work, we investigate how different RNN architectures solve the SNIPS intent detection problem. Sentences injected into trained networks can be interpreted as trajectories traversing a hidden state space. This space is constrained to a low-dimensional manifold whose dimensionality is related to the embedding and hidden layer sizes. To generate predictions, RNN steers the trajectories towards concrete regions, spatially aligned with the output layer matrix rows directions.
Underlying the system dynamics, an unexpected fixed point topology has been identified with a limited number of attractors. Our results provide new insights into the inner workings of networks that solve the intent detection task.
\end{abstract}

\end{frontmatter}

%%%%%%%%%%%%%%%%%%%%%%%%%%%%%%%%%%%%%%%%%%%%%%%%%%%%%%%%%%%%%%%%%%%%%%%%

\section{Introduction}
Modern recurrent neural networks (RNNs) are widely used to solve problems involving data sequences. Strong performance is obtained in tasks of natural language processing (NLP) such as sentiment analysis \citep{liu_sentiment_2015}, intent detection and slot filling \citep{hakkani-tur_multi-domain_2016}, and machine translation \citep{sutskever_sequence_2014}. Unfortunately, this success has not been accompanied by a deep understanding of the internal mechanisms learned by RNNs to solve concrete tasks. The exact nature of their inner workings remains an open question. The nonlinear nature of RNNs combined with high-dimensional hidden layers poses constraints on our understanding of network behavior. In addition to that, practical solutions are evolving toward increasingly complex structures \citep{bahdanau_neural_2015, sutskever_sequence_2014}. This level of complexity makes it even more challenging to understand what is happening inside the nets.   
These neural networks are used in an ever-increasing number of areas, bringing significant advances in multiple domains. Their inclusion in automated reasoning systems with a high impact on society is becoming more common. These models find regularities in the data and give predictions without explicit rules governing their behavior, the known idea of neural networks as black boxes. Therefore, it is crucial to improve our understanding of how these models solve problems and make decisions in different situations. There is an increasing societal need to develop the interpretability of neural network decision making \citep{hamon_robustness_2020}.

Several studies have tried to understand the behavior of RNNs by visualizing the activity of individual parts of the network (e.g. memory gates) during NLP tasks \citep{karpathy_visualizing_2016, strobelt_lstmvis_2018}. However, the analysis at this unit level does not provide clear interpretations. The presence of feedback connections between RNN neurons allows their interpretation as nonlinear dynamical systems \citep{martelli_introduction_1999}, opening the door to the use of a large set of well-known mathematical tools related to the theory of dynamical systems \citep{strogatz_nonlinear_2015}. Based on this idea, several works have obtained complex analytic expressions for different aspects of network dynamics \citep{haschke_input_2005, yi_convergence_2004}, such as bifurcations in the parameter space of small networks and convergence analysis. In recent years, a new reverse engineering approach has emerged for the analysis of RNNs. Instead of paying attention to the microdetails of the trained RNN behavior, a higher-level approach is considered. The state space of the trained RNN is analyzed: fixed points are located, and the dynamics of the system is linearized around them. This line of research has revealed fundamental aspects of how RNNs implement their computations \citep{sussillo_opening_2013}. These techniques have been applied to text classification problems with promising results \citep{aitken_geometry_2021,maheswaranathan_reverse_2019}. A general idea that arises is that trained networks converge to highly interpretable, low-dimensional representations associated with attractors in the RNN state space. The geometry and dimensionality of these attractors manifolds depend on the task to be solved and the internal structure of the data set.  

\subsection{Our contributions}
The main contribution of this paper is the pioneer study of the dynamics of the state space of trained RNNs for the SNIPS intent detection problem. We show that the state space is located in a manifold embedded in a low-dimensional space. The intrinsic dimensionality of this manifold is related to the size of the embedding layer and the number of neurons in the hidden layer. We also show that sentences fed into the network describe discrete trajectories through the state space toward its outer regions. A key point is the existence of distant regions from the initial states, where the trajectories end. We explain how predictions are possible due to the alignment between these peripheral areas and the directions determined by the readout matrix rows. The underlying fixed point topology of the system is obtained. Unlike sentiment analysis and document classification, we show that RNNs trained on the intent detection problem present an unexpected fixed point structure \citep{aitken_geometry_2021}.  The number of attractors and saddle points learned by the network depends on the network parameters and the type of cell. We expect to generalize these promising results from the SNIPS dataset to generic intent detection problems.

%%%%%%%%%%%%%%%%%%%%%%%%%%%%%%%%%%%%
\section{Background}
\subsection{Recurrent Neural Networks computations}

\begin{figure}[h]
\centering
\includegraphics[width=8.5cm]{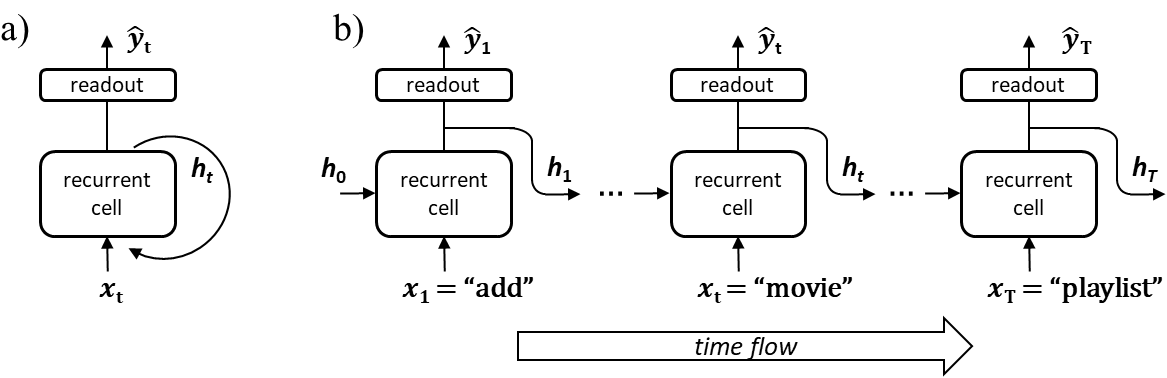}
\caption{\textbf{a)} Folded representation of RNNs emphasizing the idea of recurrence. \textbf{b)} Unfolded RNN with explicit reference to time flow.}
\label{fig:folded_unfolded}
\end{figure}

Feedforward networks (FFNs) perform a limited analysis of inputs based on the assumption of independence among the samples. However, in many situations, input samples are related in time or spatially; e.g. NLP problems are commonly studied as token sequences or the use of time series in weather forecast \citep{han_using_2021}. Some kind of memory is needed to learn the temporal information contained in the sequences. An option is to enrich the architecture of FFNs by including feedback or recurrent connections, as shown in Figure~\ref{fig:folded_unfolded}. This approach leads to the architecture called RNNs \citep{goodfellow_deep_2016}. 
%%%%%%%%%%%%%%%%%%%%%%%%%%%%%%%%%%%%

In general, computations performed by RNNs can be summarized in the following pair of difference equations:
\begin{align}
    \mathbf{h}_t = & \mathbf{F}(\mathbf{h}_{t-1},\mathbf{x}_t) \label{eq_recurrence}\\
     \mathbf{y}_t = & \mathbf{W}\mathbf{h}_t + \mathbf{b} \label{eq_readout}  
\end{align}
where $t$ is an integer index (usually representing time), $\mathbf{h}_t \in \mathbb{R}^n$ is the $n$-dimensional \textit{hidden state} of the network, $\mathbf{x}_t \in \mathbb{R}^m$ is the $m$-dimensional external input to be processed, and $\mathbf{F}$ is a nonlinear update function. The structure of $\mathbf{F}$ depends on the architecture selected to implement the recurrent cell. Given an input sequence $\mathbf{x}_1, \dots, \mathbf{x}_T$, in each computation step $t$, the network updates its state $\mathbf{h}_t$ as determined by $\mathbf{F}$, according to its previous state $\mathbf{h}_{t-1}$ and external input $\mathbf{x}_t$. To obtain the predictions of the model $\mathbf{y}_t$, these hidden states are passed through a linear \textit{readout layer} that performs an affine projection, where $\mathbf{W}$ is a $n \times n$ readout weight matrix and $\mathbf{b}$ is a bias vector. Each row $\mathbf{r}_i$ of $\mathbf{W}$ is called a readout vector. In classification problems, $\mathbf{y}_t$ consists of $N$ output logits, one for each class label. In the so-called many-to-many situations (e.g. named entity recognition), the whole stream of hidden states, $\mathbf{h}_1, \dots, \mathbf{h}_T$, is projected, obtaining a sequence of predictions $\mathbf{y}_1, \dots, \mathbf{y}_T$. In contrast, in many-to-one contexts (e.g., intent detection or sentiment analysis), a single prediction $\mathbf{y}_T$ is obtained from the last hidden state $\mathbf{h}_T$.

\subsection{Fixed points}
Systems governed by difference equations as in Equation~\ref{eq_recurrence} are called \textit{discrete-time dynamical systems} with state $\mathbf{h}_t$ at time $t$. In RNNs, the update function $\mathbf{F}$ is always nonlinear. Hence, RNNs are nonlinear discrete-time dynamical systems (NLDS) tuned to perform specific tasks. Therefore, they can be analyzed using a wide variety of well-known tools from dynamical system theory. The vector state $\mathbf{h}_t \in \mathbb{R}^n$ can be represented in a $n$-dimensional space called the \textit{state space} or the \textit{phase space} of the system, where each axis corresponds to a component of the state vector. For each initial state, the evolution of the system (governed by $\mathbf{F}$) is a sequence of states that describe a \textit{trajectory} or \textit{orbit} in the state space. The qualitative behavior of these trajectories can vary greatly between different parts of the phase space. Thus, a common line of work is to study separately different areas of the state space and the interactions between these regions. Fixed points are common points to start this analysis. A \textit{fixed point} or \textit{equilibrium point} $\mathbf{h}^*$ is a zero-motion point (or an invariant point) in the phase space. By definition, a system situated at a fixed point will remain in it. In real situations, noise or perturbations can shift the system from an equilibrium point. Therefore, it is important to understand the behavior of the system around $\mathbf{h}^*$. If the evolution of the system, when started in the neighborhood of $\mathbf{h}^*$, converges to the fixed point, $\mathbf{h}^*$ is called a \textit{stable} fixed point. On the contrary, if the system diverges from it, the fixed point is called \textit{unstable}. Finally, a \textit{saddle point} behaves as a stable equilibrium point for some trajectories and as an unstable point for others.

\subsection{Linearization}
Fixed points have an important property; the Hartman-Grobman theorem \citep{hartman_ordinary_2002} asserts that the behavior of an NLDS around a fixed point $\mathbf{h}^*$ has the same qualitative behavior\footnote{Only valid for hyperbolic fixed points. An interested reader can find more information at \citep{hartman_ordinary_2002}.} as its linearization $\mathbf{JF}(\mathbf{h}^*)$ (that is, the Jacobian of $\mathbf{F}$ around $\mathbf{h}^*$). Therefore, NLDS analysis is performed in two steps: a) to identify the fixed points of the system and b) to analyze their linearized behavior. Linearized systems analysis is much easier and involves the decomposition of the Jacobian $\mathbf{J = R \Lambda L}$ where $\mathbf{\Lambda}$ is a diagonal matrix such that the elements $\lambda_i$ are the $n$ eigenvalues of an $n$-dimensional linear system and $\mathbf{L} = \mathbf{R}^{-1}$. Each $\lambda_i$ has an associated eigenvector $\mathbf{r}_i$, row of the matrix $\mathbf{R}$.  From this matrix decomposition, the state of a $n$-dimensional linear dynamical system can be interpreted as the linear composition of $n$ independent one-dimensional exponential dynamics (also called \textit{modes} or patterns of activity). Each of them takes place along the direction of the state space given by the eigenvector $\mathbf{v}_i$. These directions are invariant lines in the phase space. Therefore, the motion near $\mathbf{h}^*$ can be obtained as the linear composition of these $n$ one-dimensional systems. The behavior of a mode along the direction $\mathbf{v}_i$ is controlled by its associated eigenvalue $\lambda_i$ given by $\lambda_i^tb_i$. If $|\lambda_i| > 1$, the component $\mathbf{h}_t$ in direction $\mathbf{v}_i$ grows exponentially. On the contrary, if $|\lambda_i| < 1$, it shrinks exponentially. Therefore, the stability of $\mathbf{h}^*$ is determined by the spectral radius of $\mathbf{JF}(\mathbf{h}^*)$. If all $\lambda_i$ are within the unit circle, then $\mathbf{h}^*$ is a stable fixed point. In a saddle point, some eigenvalue has a norm greater than 1.
On the other hand, if every $\lambda_i$ is beyond the unit circle, then $\mathbf{h}^*$ is a totally unstable equilibrium point.

\subsection{Basins of attraction and saddle points}
In many common situations, systems without external input evolve toward certain regions of the state space; such a converging point or region is called an \textit{attractor}. Stable fixed points are the simplest attractors. The \textit{basin of attraction} of an attractor is the region of the phase space (i.e. the set of all initial states) from which the system evolves towards the attractor. Any initial condition in that region will be iterated in the attractor. The whole state space is divided into basins of attraction associated with the attractors. In this partition, saddle points play a key role in controlling the interaction between attractors. Saddle points mostly have stable modes (i.e. modes associated to $\lambda_i < 1$), also called \textit{stable manifolds} with only a small set of unstable modes (associated to $\lambda_i > 1$), also known as \textit{unstable manifolds}. A region of state space can be funneled through its many stable modes and then sent to different attractors forced by an unstable mode. As a result of these state-space management operations, a stable saddle point manifold becomes a frontier between the basins of attraction \citep{ceni_interpreting_2020}. The \textit{index} of a saddle point is defined as the number of unstable manifolds of the fixed point.

\subsection{Reverse engineering RNNs of classification tasks}
RNNs, as NLDS, can be analyzed with tools of dynamical system theory. Modern RNN architectures are made up of hidden layers with hundreds of neurons, which implies high-dimensional hidden states. Traditional dynamical system analysis is performed considering individual neurons as parameters of the system. This high dimensionality of RNNs makes a standard analysis of state spaces difficult. A recent line of work considers the computational mechanisms learned by RNNs from a higher-level perspective \citep{sussillo_opening_2013}. These reverse engineering techniques are applied to analyze the dynamics of networks trained for specific tasks. The behavior of an RNN can be inferred from the structure of its state space: the fixed points, their linearized dynamics, and the interactions between these equilibrium points. For example, binary sentiment analysis and text classification problems share a common underlying mechanism \citep{aitken_geometry_2021}. In solving the task, their hidden state trajectories lie largely in a low-dimensional subspace of the full state space. An attractor manifold lies in this subspace that accumulates evidence (that is, keeps track) for each class as they process tokens of the text. The concrete dimensionality and geometry of this attractor manifold are determined by the structure of the dataset. In binary sentiment classification tasks, hidden states move along a line of stable fixed points \citep{maheswaranathan_reverse_2019}. In general, the attractors of a categorical classification of $N$ classes form a $(N-1)$ dimensional simplex \citep{aitken_geometry_2021}. This dimensionality reflects the number of scalar quantities that the network remembers to classify.

\subsection{Intent detection problem}
Spoken Language Understanding (SLU) \cite{qin2021survey} is an important element of many natural language tools, such as dialogue systems.  Its role is to capture the semantics of user utterances for use in other processes (i.e. 
question-answering or dialogue management). Three key tasks are involved in building this semantic frame: domain classification, intent detection, and slot filling, as shown in Table~\ref{tab:utterance_example}  
\citep{tur_spoken_2011}. The intent is the speaker's desired outcome from the utterance. Only if the user's intent is clearly identified can the query be routed to the correct subsystem.

\begin{table}[h]
\caption{Example of utterance, its domain, intent and slots.}
\centering
\renewcommand{\arraystretch}{1.5}
\begin{tabular}{|l|c|c|c|c|c|c|} 

\hline
\textbf{query} & find & recent & comedies & by & James & Cameron \\
\cline{1-7}
\textbf{slots} &  O & B-date & B-genre & O & B-dir & I-dir \\
\cline{1-7}
\textbf{intent} & \multicolumn{6}{c|}{find\_movie}  \\
\cline{1-7}
\textbf{domain} & \multicolumn{6}{c|}{movies}  \\
\hline
\end{tabular}
\label{tab:utterance_example}
\end{table}

Two datasets have been widely used as benchmarks for intent detection models. First, SNIPS \citep{coucke_snips_2018} is a balanced dataset with 7 intents, designed in the context of English voice assistants. On the contrary, ATIS \cite{hemphill_atis_1990} is a heavily imbalanced 26-intent dataset, which contains real conversations with English-speaking customers who request information about flights. Some of its utterances are labeled with more than one intent. Recently, a new multilingual MASSIVE dataset \citep{fitzgerald_massive_2022} was released. In this case, the original 60-intent SLURP dataset \citep{bastianelli_slurp_2020} has been localized in 51 different languages. The number of domains, intents, and slots in each of these datasets is shown in Table~\ref{tab:datasets_comparison}.
\begin{table}[h]
\caption{Intent detection datasets comparison.}
\centering
\begin{tabular}{lccccc} 
\toprule
Name & Langs & Utters per lang & Domains & Intents & Slots \\
\midrule
SNIPS & 1 & 14484 & - & 7 & 53 \\
ATIS &  1 & 5871 & 1 & 26 & 129 \\
MASSIVE & 51 & 19521 & 18 & 60 & 55 \\
\bottomrule
\end{tabular}
\label{tab:datasets_comparison}
\end{table}

The degree of imbalance of each dataset is shown in Table~\ref{tab:datasets_imbalance}. The ATIS dataset is extremely imbalanced, with almost 74\% of the samples belonging to a single intent. The MASSIVE dataset presents a large number of intents combined with a certain degree of imbalance. In this paper, we focus on the SNIPS dataset, trying to avoid possible side effects due to imbalance or a high number of intents.
\begin{table}[h]
\caption{Degree of imbalance of the different datasets.}
\centering
\begin{tabular}{lcccc} 
\toprule
Name & \CellWithForceBreak{\#samples in\\ larger intent} & \CellWithForceBreak{larger \\ intent (\%)} & \CellWithForceBreak{\#samples in \\ smaller intent} & \CellWithForceBreak{smaller \\ intent (\%)} \\
\midrule
SNIPS & 2100 & 14.5 & 2042 & 14.1 \\
ATIS &  4298 & 73.7 & 1 & 0.02\\
MASSIVE & 1190 & 6.9 & 6 & 0.04\\
\bottomrule
\end{tabular}
\label{tab:datasets_imbalance}
\end{table}

Intent detection is usually approached as a supervised classification task that associates the entire input sentence with a label (or intent) of a finite set of classes \citep{liu_review_2019}. Given the ability of RNNs to capture temporal dependencies, RNNs have been widely used to solve intent detection problems \citep{ravuri_recurrent_2015}. 

\section{Experiments}
We have divided our analysis into four steps: a) train different RNN architectures to solve the intent detection problem for the SNIPS dataset; b) obtain the state space learned by the RNN; c) analyze the manifold structure in which the state space is embedded; and d) obtain the structure of fixed points that underlies these trained RNNs.

We use TensorFlow 2 \citep{abadi_tensorflow_2016} to train basic RNNs with a trainable embedding layer with \textit{embed\_dim} neurons (i.e. no pretrained embeddings have been considered), a unidirectional RNN layer with \textit{hidden\_dim} neurons, and a final dense layer. Tokenization was implemented using a TextVectorization layer with a vocabulary truncated to 1K words (from a total of 10.5K words). Three different implementations of recurrent cells were considered: standard (or vanilla), LSTM \citep{hochreiter_long_1997}, and GRU \citep{cho_learning_2014}. RNNs have been trained on the SNIPS \citep{coucke_snips_2018} intent detection datasets. The optimized loss function is the usual one for classification problems (i.e. multiclass cross-entropy). Network training was carried out using an Adam optimizer \citep{kingma_adam_2015} with a batch size of 32, and a learning rate $\eta=$ 5e-4. The number of training epochs was determined using early stopping \citep{prechelt_early_1996} with a patience of 2 epochs. No additional hyperparameters were tuned, using the default value for the rest of the parameters. During training, no dropout \citep{srivastava2014dropout} or other regularization techniques have been used. For each architecture, we selected the best performing network based on a validation dataset. These validation subsets were obtained as a random sample 20\% from the training dataset. Given the balanced distribution of the SNIPS target class, we evaluated the performance of the model by computing the accuracy in a test dataset not used during the training process. Dataset was randomly divided 80/20 to obtain the train and test subsets, respectively.

\section{Results}
\subsection{Intent detection low-dimensional dynamics}\label{section: low_dim_dynamics}
\begin{figure}[h]
\centering
\includegraphics[width=8.3cm]{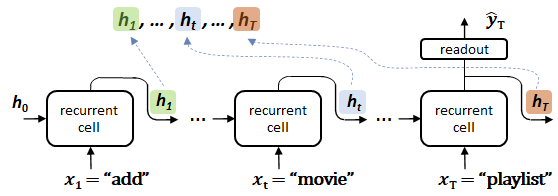}
\caption{Sequence of hidden states associated to a tokenized input sentence fed into a RNN.}
\label{fig:hidden_states_inspection}
\end{figure}

In this section we show that the state space learned by an RNN is constraint to a low-dimensional hypersurface (or manifold). Before sentences are injected into RNNs, a tokenization mechanism is needed to transform each natural language phrase into a sequence $\mathbf{x}_1, \dots, \mathbf{x}_T$, where $\mathbf{x}_i \in \mathbb{R}^m$ is a $m$-dimensional vector and $T$ is the number of tokens in the sentence \citep{jurafsky_speech_2021}. As shown in Figure~\ref{fig:hidden_states_inspection}, the sequential injection of these tokens generates a sequence of activations $\mathbf{h}_1, \dots, \mathbf{h}_T$ in the output of the hidden layer. Each hidden state $\mathbf{h}_i \in \mathbb{R}^n$ is a $n$-dimensional vector (with $n = hidden\_dim$) given by Equation~\ref{eq_recurrence}. The set of hidden states visited by sentences injected into a network is called the state space learned by the trained RNN. 

Under the assumption that the discrete points of the state space can be summarized by a manifold embedded in a higher-dimensional space, the next natural question is to determine its intrinsic dimensionality. Different measures of this dimensionality have been proposed \citep{levina_maximum_2004, camastra_estimating_2002}. However, related works suggest that the variance explained threshold is the one that best fits the empirical data in classification tasks \citep{aitken_geometry_2021}. This measure considers the number of dimensions of the principal component analysis (PCA) \citep{jolliffe_principal_2002} needed to reach a certain percentage of the variance explained. This threshold is typically set at a fixed value 95\%. 

\begin{figure}[h]
\centering
\includegraphics[width=8.5cm]{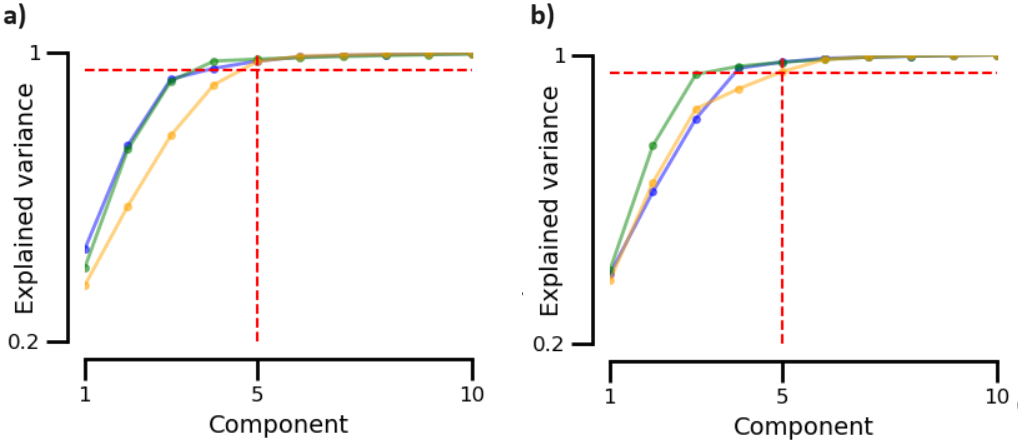}
\caption{Variance explained of visited states vs principal components of a RNN trained on the SNIPS dataset. \textbf{a)} GRU(e:16,h:16). \textbf{b)} GRU(e:10,h:10).}
\label{fig:explained_variance}
\end{figure}

Following this procedure, all sentences from the SNIPS test dataset were injected into a trained RNN. The state space points were concatenated, and by performing a principal component analysis, the variance captured by each principal component was obtained. In Figure~\ref{fig:explained_variance} the accumulated explained variance versus the number of principal components is shown for two combinations of $hidden\_dim$ and $embed\_dim$. Variances for the vanilla, LSTM, and GRU cell types are represented in green, yellow, and blue, respectively. The horizontal dashed red line indicates the variance explained threshold at 95\%. The vertical red line shows the number of principal components needed to surpass this threshold, i.e. the (intrinsic) dimensionality $d$ of the state space. For simplicity, we denote by \textit{cell\_type(e:x,h:y)} an RNN with \textit{cell\_type} recurrent unit, \textit{x} neurons in the embedding layer and a hidden layer of size \textit{y}

\begin{figure}[h]
\centering
\includegraphics[width=8.75cm]{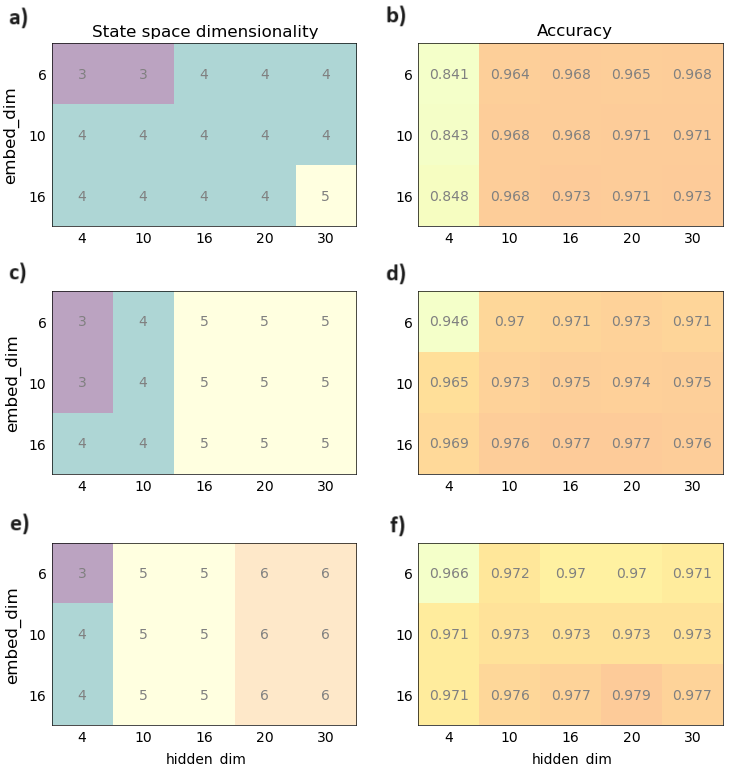}
\caption{State space dimensionality and accuracy of RNNs trained on the SNIPS dataset for different embed\_dim and hidden\_dim combinations. \textbf{(a,b)}: Vanilla cell. \textbf{(c,d)}: GRU cell. \textbf{(e,f)}: LSTM cell}
\label{fig:dimensionality_greed_search}
\end{figure}

From related work, the state space dimensionality of RNNs solving categorical text classification problems is $N-1 \ll hidden\_dim$, with $N$ the number of classes \citep{aitken_geometry_2021}. According to these proposals, the expected dimensionality $d_{e}$ of the 7-class SNIPS dataset must be $N-1 = 6$. We analyze whether this assertion also holds for the intent detection problem. A comparative analysis of the dimensionality of the state space and the accuracy of RNNs trained on the SNIPS dataset is shown in Figure~\ref{fig:dimensionality_greed_search}.

Different combinations of embedding and hidden layer sizes have been tested. The values presented correspond to the average accuracy and median state space dimensionality of ten trains with different seeds for each pair \textit{(hidden\_dim, embed\_dim)}. Contrary to our initial expectations, the dimensionality of the state space depends on both network design parameters. Furthermore, at least in the range of tested parameters, for any \textit{embed\_dim}, a \textit{hidden\_dim} could be found such that it solves the problem with a state space manifold of dimensionality $d < d_{e} \ll n\_hidden$ (without accuracy degradation).

\subsection{Intent detection state space projection}
\begin{figure}[h]
\centering
\includegraphics[width=8.5cm]{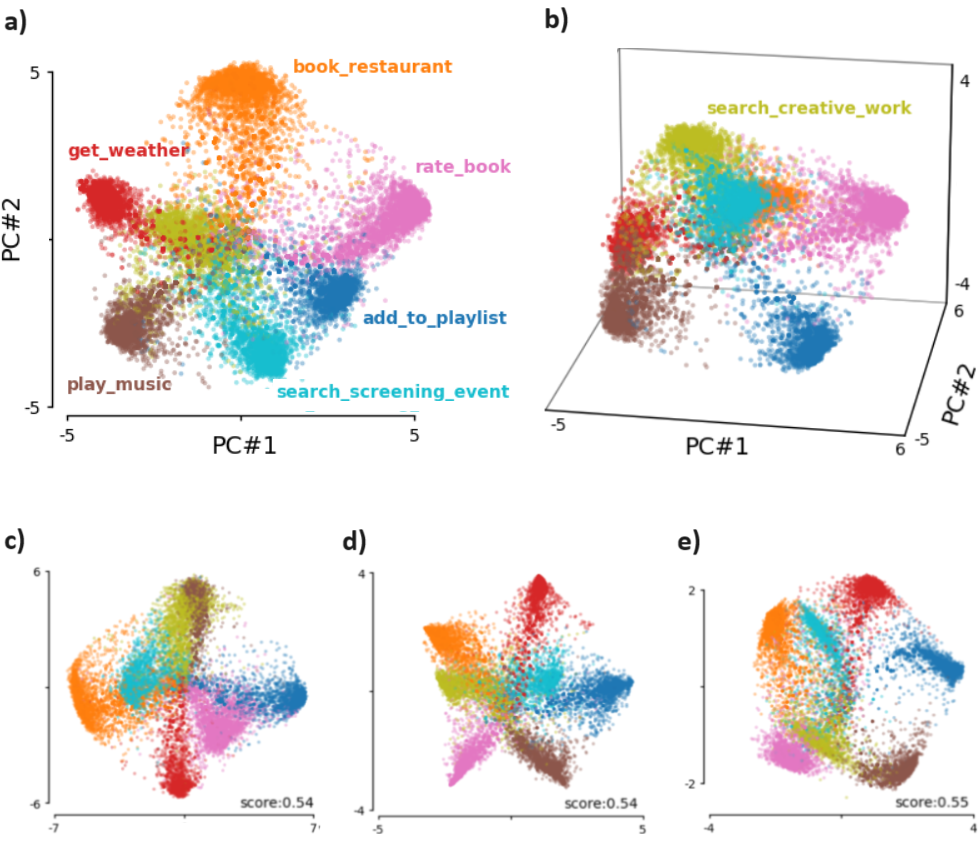}
\caption{State space top-2 and top-3 PCA projections of RNNs trained on the SNIPS dataset. \textbf{a)} GRU(e:16,h:16). \textbf{b)} GRU(e:16,h:16). \textbf{c)} Vanilla(e:20,h:20). \textbf{d)} LSTM(e:10,h:10). \textbf{e)} GRU(e:10,h:10).}
\label{fig:projected_state_space}
\end{figure}

In the following sections, we analyze the spatial arrangement of hidden states in trained RNNs, taking advantage of the low dimensionality ($d << n\_hidden$) of their state spaces. For that, given a state space, it can be projected in the linear subspace given by the k-top principal components of Section \ref{section: low_dim_dynamics}. For this, a linear transformation $\mathbf{U}$ is applied to each hidden state:
\begin{equation}
\mathbf{p}_i = \mathbf{h}_i\mathbf{U} 
\label{eq:projection}
\end{equation}

\noindent where $\mathbf{h}_i \in \mathbb{R}^n$ is a $n$-dimensional hidden vector, $\mathbf{p}_i \in {R}^k$ is the $k$-dimensional projected hidden state and $\mathbf{U}$ is a $n \times k$ projection matrix. Top-2 and top-3 projections are generally considered for visualization purposes; meanwhile, higher-dimensional projections are useful for dimensionality analysis (as in Section \ref{section: low_dim_dynamics}). In Figures~\ref{fig:projected_state_space}a) and b), 2D and 3D projections of the state space learned by a GRU(e:16,h:16) are shown. 
Each state is colored according to the intent of its source sentence. In both projections, hidden states appear seemingly grouped on the basis of their associated intent.

To numerically verify the presence of these clusters, we have applied a classical KMeans \citep{lloyd_least_1982} clustering method with 7 clusters, random initial centroids, and Euclidean distance metric. The resulting state space partition was evaluated against the true labels, with a silhouette technique \citep{rousseeuw_silhouettes_1987}. The silhouette coefficient of a point measures its intra-cluster and the nearest-cluster distances to the rest of points. The silhouette coefficients take values in the range [-1, 1]. A coefficient near +1 indicates that the point is far from neighboring clusters. Points close to the decision boundary between two neighboring clusters present values around 0. Finally, negative values indicate that the sample might have been assigned to a wrong cluster. The silhouette score is obtained by computing the mean silhouette coefficient on all samples. The score threshold, which is used to assess the quality of a cluster, is commonly set at 0.5. A score above 0.5 indicates a high-quality cluster. 

\begin{figure}[h]
\centering
\includegraphics[width=8.5cm]{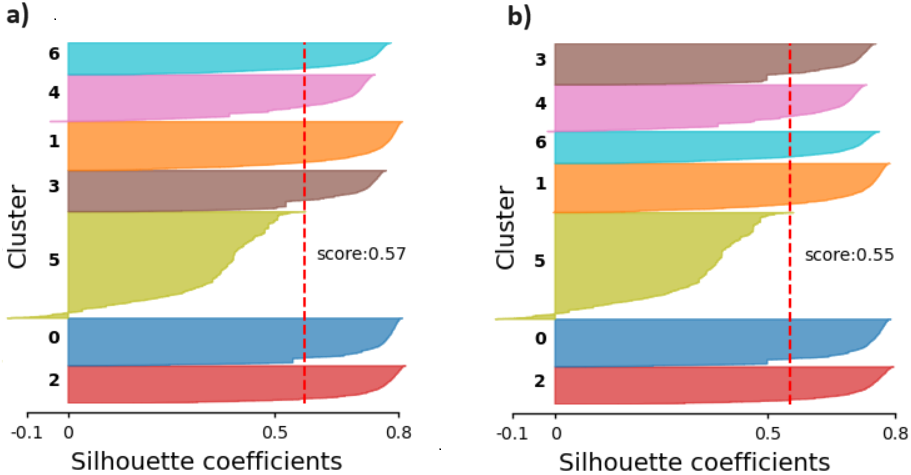}
\caption{Silhouette evaluation of a GRU(e:16,h:16) state space KMeans partition. \textbf{a)} Distances obtained considering only top-5 PCA projections. \textbf{b)} 16-dimensional hidden states considered for distances.}
\label{fig:silhouette}
\end{figure}

Figure~\ref{fig:silhouette} shows the silhouette scores for a trained GRU(e:16,h:16) in two different situations. In Figure~\ref{fig:silhouette}a) the distances between the points were calculated considering only the top-5 projections of the hidden states (since the intrinsic dimensionality of this RNN configuration is $d=5$ from Section \ref{section: low_dim_dynamics}). On the other hand, in Figure~\ref{fig:silhouette}b) fully 16-dimensional hidden states were considered. In both cases, the resulting partitions have similar global scores (dashed red line) above the 0.5 threshold. This comparison confirms the utility of analyzing the state spaces manifold considering uniquely the top-d projections, with $d$ the intrinsic dimensionality of the state space.
The partition of the state space in different regions associated with intents is independent of the type of cell, $hidden\_dim$ and $embed\_dim$. In Figures ~\ref{fig:projected_state_space}a), b) and c) the projections of the top-2 state space projections of different networks are shown for a Vanilla(e:20,h:20), a LSTM(e:10,h:10), and a GRU(e:10,h:10). For all configurations, the silhouette scores associated with a KMeans partition of the state space confirm the presence of clusters of states.

\subsection{Sentences trajectories}\label{sentences_trajectories}
In this section, we study how sentences move through state space when injected into a trained RNN. In practice, the state space of an RNN is obtained by feeding sentences to the network and aggregating the resulting visited states. In response to each input $\mathbf{x}_i$ the current state $\mathbf{h}_t$ is updated according to Equation~\ref{eq_recurrence} obtaining the next state $\mathbf{h}_{t+1}$. Hence, an input sequence $\mathbf{x}_1, \dots, \mathbf{x}_T$, when injected into the network, produces a new sequence $\mathbf{h}_1, \dots, \mathbf{h}_T$ that describes a trajectory traversing the state space.  The points of the trajectory can be projected onto the principal components of the low-dimensional state space using Equation~\ref{eq:projection}. 

\begin{figure}[h]
\centering
\includegraphics[width=8.5cm]{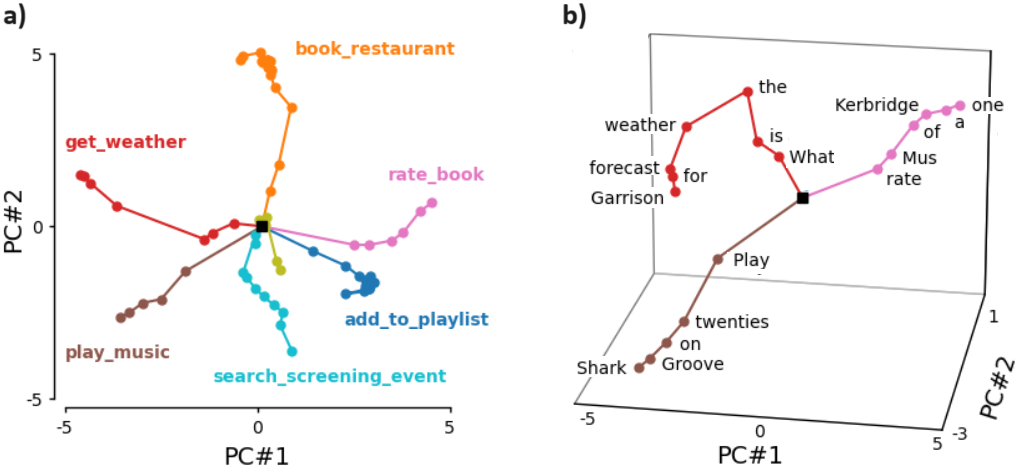}
\caption{Trajectories of example sentences projected on the state space of a GRU(e:16,h:16) trained on the SNIPS. \textbf{a)} Top-2 PCA projections. \textbf{b)} Top-3 PCA projections.}
\label{fig:projected_sentences}
\end{figure} 

Figure~\ref{fig:projected_sentences}a) presents the 2D projection of the trajectories associated with an example sentence for each intent. Each $\mathbf{h}_i$ has been highlighted with bullets. The initial state $\mathbf{h_0}$ of the recurrent cells is marked by black squares, and lines joining the states were added to emphasize the sequence of movements. In Figure~\ref{fig:projected_sentences}b) details of three sentences are shown. In red, with "\textit{get\_weather}" intent: "\textit{What is the weather forecast for Garrison}". In brown, with "\textit{play\_music}" intent; "\textit{Play twenties on Groove Shark}" and, with intent \textit{"rate\_book"}, in purple: "\textit{Rate Mus of Kerbridge a one}"). The input tokens are indicated next to the hidden state generated. As tokens are received, the orbits diverge from the origin moving towards concrete outer areas of the state space.

\subsection{Model inference mechanism}
In this section, we analyze the regions of the state space toward which RNNs direct the trajectories.  In intent detection problems, the final hidden state of each sentence plays a crucial role. As shown in Figure~ \ref{fig:hidden_states_inspection}, for prediction purposes, intermediate states are discarded, and only the last hidden state is considered. Given a sentence, the logits of each class are obtained projecting the last hidden state $\mathbf{h}_T$ onto each of the readout vectors $\mathbf{r}_i$ (that is, the rows of the readout matrix $\mathbf{W}$) as follows:
\begin{equation}
\mathbf{y} = \mathbf{y}_T = \mathbf{W}\mathbf{h}_T = [\mathbf{r}_1|\ldots|\mathbf{r}_n]^T\mathbf{h}_T
\end{equation}

\begin{figure}[h]
\centering
\includegraphics[width=8.5cm]{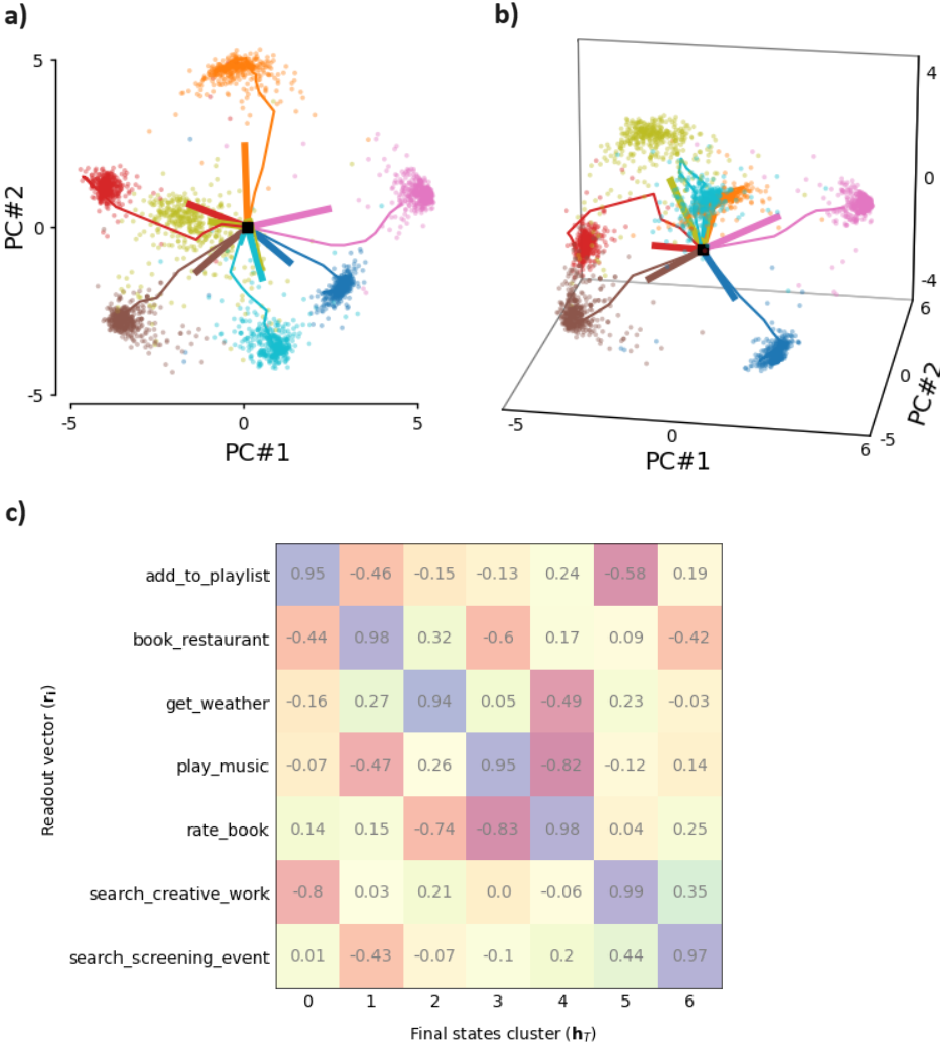}
\caption{Final hidden states and readout vectors of a GRU(e:16,h:16). An example trajectory for each intent. \textbf{a)} Top-2 PCA projections. \textbf{b)} Top-3 PCA projections. \textbf{c)} Cosine similarity between clusters of final states and readouts.} 
\label{fig:annotated_and_readouts}
\end{figure}

The index $i$ with the highest scalar value $\mathbf{r}_i^T\mathbf{h}_T$ will be output as the predicted intent. In Figures~\ref{fig:annotated_and_readouts}a) and b), the last hidden states $\mathbf{h}_T$ of each sentence in the SNIPS test dataset are projected onto its principal components. The states are colored according to the intent of the source sentence. The readout vectors $\mathbf{r_i}$ can be interpreted as directions that can also be projected in the state space. $\mathbf{r_i}$'s are represented following the same color schema.  For illustration purposes, some trajectories have been added. We hypothesize that trajectory of sentences evolve through the low-dimensional state space towards concrete peripheral areas distant from the initial states.

We performed a 7-cluster KMeans analysis on the set composed of the final state of each sentence in the SNIPS test dataset. The results for different RNN configurations are presented in Table \ref{tab:last_states_cluster_analysis}. In all cases, silhouette scores greater than 0.75 clearly confirm the presence of as many clusters of final states as intents. For each configuration, the distances $ds = \{ds_1,\ldots,ds_7\}$ between the centroid of each cluster and the initial state $\mathbf{h_0}$ were calculated. In all the configurations tested, the standard deviation of the distances $\sigma(ds)$ is much lower than its mean value $\bar{ds}$. These results suggest that centroids are located in positions roughly equidistant from the initial state $\mathbf{h_0}$. 

\begin{table}[h]
\caption{Clusters of final states for different RNN configurations: silhouette scores, statistics of distances centroid - $\mathbf{h_0}$ and alignment data.}
\centering 
  \setlength{\tabcolsep}{4.0pt}
  \begin{tabular}{l c c c c wc{1em} wc{1em}}
    \toprule
      Cell &
      embed &
      hidden &
      Silhouette &
      Alignment &
      \multicolumn{2}{c}{\CellWithForceBreak{Distances}}  \\
      type & dim & dim & score & mean & $\bar{ds}$ & $\sigma(ds)$ \\
    \midrule
    Vanilla & 20 & 20 & 0.75 & 0.957 & 5.38 & 0.30 \\
    GRU & 16 & 16 & 0.80 & 0.964 & 4.81 & 0.34 \\
    LSTM & 10 & 10 & 0.81 & 0.963 & 4.09 & 0.73 \\
    \bottomrule
  \end{tabular}
\label{tab:last_states_cluster_analysis}
\end{table}

Given a sentence with true intent $I$ and final state $\mathbf{h_T}$, to correctly generate a prediction, the trained RNN must ensure that the value of $\mathbf{r}_i^T\mathbf{h}_T$ is greater for $i = I$ than for any other $i \ne I$. To maximize this dot product, for each intent $I$ the readout vector $\mathbf{r_I}$ must be as aligned as possible with the final states $\{\mathbf{h_T\}_I}$ of the cluster associated with the intent $I$. Cosine similarity is widely used to measure the degree of alignment between two vectors \citep{dangeti_statistics_2017}. It computes the cosine of the angle between both vectors, taking values in the range [-1,1]. Similarity 1 indicates that both vectors are perfectly aligned, pointing in the same direction. Two orthogonal vectors present value 0. Similarities close to -1 indicate that the vectors are aligned but point in opposite directions. In Figure~\ref{fig:annotated_and_readouts}c) is shown the cosine similarity between all pairs ($r_i$, $centroid_j$) for a trained GRU (e: 16, h: 16). As a result, each $r_i$ has a single almost perfectly aligned centroid, with similarity values greater than 0.9. In Table~\ref{tab:last_states_cluster_analysis} we have computed the mean value of the distances between the aligned pairs $r\_i, centroid\_j$ for different configurations.  

\subsection{Fixed point structure}
In the previous section we have shown that sentences traverse a low-dimensional state space in a journey steered by the RNN towards final state clusters almost equidistant from the initial state. Each of these clusters is aligned with a paired readout vector to maximize its dot product. This dynamic suggests the existence of an underlying fixed point structure. 

In general, we can write the first-order approximation to the RNN dynamics \citep{khalil_nonlinear_2013} around an expansion point $(\mathbf{h}_e, \mathbf{x}_e)$ as:
\begin{equation}
\mathbf{h}_t \approx \mathbf{F}(\mathbf{h}_e,\mathbf{x}_e) + 
\mathbf{J}^{rec}\mathbf{F}|_{(\mathbf{h}_e,\mathbf{x}_e)} \Delta \mathbf{h}_{t-1} + \mathbf{J}^{inp}\mathbf{F}|_{(\mathbf{h}_e,\mathbf{x}_e)} \Delta \mathbf{x}_t 
\label{eq:expansion_rnn}
\end{equation}
where $\Delta \mathbf{h}_{t-1} =  \mathbf{h}_{t-1} - \mathbf{h}_e$, $\Delta \mathbf{x}_t =  \mathbf{x}_t - \mathbf{x}_e$ and \{$\mathbf{J}^{rec}\mathbf{F}, \mathbf{J}^{inp}\mathbf{F}\}$ correspond to the Jacobian matrices of the update function $\mathbf{F}$ particularized at the expansion point. In concrete, the \textit{recurrent Jacobian} $\mathbf{J}^{rec}\mathbf{F}$ captures the local dynamics associated with the recurrence, and the \textit{input Jacobian} $\mathbf{J}^{inp}\mathbf{F}$ represents the sensitivity of the system to input tokens, with: 
\begin{equation}
    J_{ij}^{rec}\mathbf{F} = \frac{\partial{F_i}}{\partial{h_j}}\Bigr|_{(\mathbf{h}_e,\mathbf{x}_e)} \qquad J_{ij}^{inp}\mathbf{F} = \frac{\partial{F_i}}{\partial{x_j}}\Bigr|_{(\mathbf{h}_e,\mathbf{x}_e)}
\end{equation}
When the expansion is performed in a fixed point, then $\mathbf{h}^* = \mathbf{F}(\mathbf{h}^*,\mathbf{x})$ and Equation~\ref{eq:expansion_rnn} simplifies to a linear dynamical system equation:
\begin{equation}
\Delta \mathbf{h}_t  = \mathbf{h}_t - \mathbf{h}_e \approx
\mathbf{J}^{rec}\mathbf{F}|_{(\mathbf{h}^*,\mathbf{x}^*)} \Delta \mathbf{h}_{t-1} + \mathbf{J}^{inp}\mathbf{F}|_{(\mathbf{h}^*,\mathbf{x}^*)} \Delta \mathbf{x}_t 
\label{eq:expansion_rnn_simplified}
\end{equation}
\noindent Due to the presence of time-dependent inputs $\mathbf{x_t}$ (that is, the tokens associated with sentences injected into the network), RNNs are nonautonomous dynamical systems. Analysis of nonautonomous systems requires sophisticated mathematical tools. 
For this reason, the reverse engineering point of view attempts to explain the behavior of the system in three steps: a) the obtention of the topological structure of the fixed points for constant input $\mathbf{x}_t$, b) the analysis of the linearized system around these fixed points under constant input, and c) the study of how the linearized behavior is altered (a.k.a. deflected) under the influence of nonconstant external inputs \citep{aitken_geometry_2021}.

In this section, we focus on the first step, the identification of the fixed point structure. This approach starts with locating the set of points $\{\mathbf{h}_1^*, \mathbf{h}_2^*, \ldots\}$ in the state space such that $\mathbf{h}_i^* = \mathbf{F}(\mathbf{h}_i^*, \mathbf{x})$ where $\mathbf{x}$ is a constant input to the system. 
Typically, $\mathbf{x}^*$ is set to $\mathbf{0}$, with a clear physical meaning; given an initial state, the system on its own evolves, describing a trajectory without external energy injected into the system. In related work, not only are truly fixed points considered, but also very slow motion points \citep{sussillo_opening_2013}. In the following, we use the term fixed point to include not only classical fixed points but also these approximate fixed points, that is, $\mathbf{h}_i^* \approx \mathbf{F}(\mathbf{h}_i^*,\mathbf{0})$. 

Different numerical procedures can be considered to identify fixed points \citep{katz_using_2018, sussillo_opening_2013}. The speed of a point in a state space can be characterized as the magnitude $q$ of the displacement generated by the update function $\mathbf{F}$ applied at the point $\mathbf{h}$ :
\begin{equation}
q = \frac{1}{2} \|\mathbf{h}-\mathbf{F}(\mathbf{h},\mathbf{0})\|_2^2    
\end{equation}
We found slow motion (and zero-motion) points by performing a numerical optimization process that minimizes this quantity $q$ \citep{golub_fixedpointfinder_2018}.
Typically, state spaces have multiple regions with different behavior. For this reason, the optimization procedure must be run multiple times with different initial conditions (ICs). In our case, more than 25K initial states were considered extracted from the trajectories visited by the sentences in the SNIPS test dataset. The critical points of the loss function with a speed value $q < 10^{-8}$ were considered slow motion or approximate fixed points. Under the assumption of $\mathbf{x} = \mathbf{0}$, Equation~\ref{eq:expansion_rnn_simplified} simplifies to:
\begin{equation}
\Delta \mathbf{h}_t  = \mathbf{h}_t - \mathbf{h}_e \approx
\mathbf{J}^{rec}\mathbf{F}|_{(\mathbf{h}^*,\mathbf{0})} \Delta \mathbf{h}_{t-1}
\label{eq:expansion_rnn_simplified_x0}
\end{equation}
The stability of each approximate fixed point $\mathbf{h^*
_i}$ is determined by the eigenvalues of  $\mathbf{J}^{rec}\mathbf{F}|_{(\mathbf{h}_i^*,\mathbf{0})}$, the Jacobian evaluated at that point. In Table \ref{tab:fps_identified} the fixed points of different RNNs trained in the 7-class SNIPS dataset are presented. The number of intents involved in the problem acts as an upper bound of the number of attractors. In all cases, a group of saddle points is identified. The amount of critical points learned by the network depends on the type of cell and the size of the hidden layer and the embedding. The presence of saddle points with an index higher than one suggests the existence of a hierarchy of critical points, visited by the trajectories. 

\begin{table}[h]
\caption{Approximated fixed points ($q < 10^{-8}$) of RNNs trained on the SNIPS dataset.}
\centering
\begin{tabular}{lccccc} 
\toprule
    Cell &
    embed &
    hidden &
    \#stable &
    \multicolumn{2}{c}{\#saddle points}  \\
      type & dim & dim & points & 1-index & higher-index \\
\midrule
Vanilla & 10 & 10 & 5 & 9 & 7\\
Vanilla & 16 & 16 & 4 & 5 & 4\\
GRU     & 10 & 10 & 5 & 6 & 1\\
GRU     & 16 & 16 & 3 & 3 & 1\\
\bottomrule
\end{tabular}
\label{tab:fps_identified}
\end{table}

We analyze the location of the critical points projected in the top-3 components of the state space. The distances $r_s$, $r_1$, and $r_2$ (for stable, 1-index and 2-index, respectively) of each type of fixed points to the origin were obtained. The mean distances $\bar{r}$ and the standard deviations $\sigma(r)$ for different network configurations are presented in Table~\ref{tab:fps_location}.

\begin{table}[h]
\caption{Mean and standard deviation (in the projected state space) of distances to origin of attractors ($r_s$), 1-index ($r_1$) and 2-index saddles ($r_2$).}
\centering
\begin{tabular}{wl{2.25em} wc{2em} wc{2em} wc{1.3em} wc{1.3em} wc{1.7em} wc{1.7em} wc{1.7em} wc{1.7em}}
\toprule
    Cell &
    embed &
    hidden &
    \multicolumn{2}{c}{stable points}  &
    \multicolumn{2}{c}{1-index}  &
    \multicolumn{2}{c}{2-index}  \\
    type & dim & dim & $\bar{r}_s$ & $\sigma(r_s)$
         & $\bar{r}_1$ & $\sigma(r_1)$ & $\bar{r}_2$ & $\sigma(r_2)$\\
\midrule
Vanilla & 10 & 10 & 3.93 & 0.10 & 3.52 & 0.13 & 3.07 & 0.31 \\
Vanilla & 16 & 16 & 4.73 & 0.29 & 4.25 & 0.38 & 4.03 & 0.36 \\
GRU & 10 & 10 & 3.72 & 0.32 & 3.43 & 0.33 & 1.72 & 0 \\
GRU & 16 & 16 & 4.71 & 0.15 & 4.27 & 0.18 & 3.29 & 0 \\
\bottomrule
\end{tabular}
\label{tab:fps_location}
\end{table}

\begin{figure}[h]
\centering
\includegraphics[width=8.5cm]{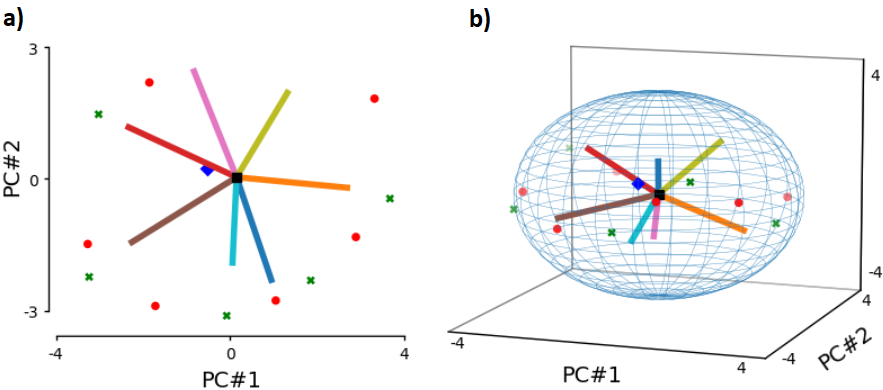}
\caption{Attractors (green), 1-index saddle points (red), 2-index saddle point (blue) and readout vectors of a GRU(e:10,h:10) trained in the SNIPS dataset. \textbf{a)} Top-2 PCA projections. \textbf{b)} Top-3 PCA projections.}
\label{fig:fixed_points_2d_3d}
\end{figure}

Each family of fixed points can be located on the surface of a hypersphere of radius $r_x$ with $x={e,1,2}$. This radius depends on the dimension of the embedding and the hidden layer. In Figure~\ref{fig:fixed_points_2d_3d} is shown the spatial arrangement of the critical points associated with a GRU(e:10,h:10). Each pair of attractors (in green) is separated by a 1-index saddle point (in red). The unique 2-index saddle point is marked in blue. 

\section{Conclusion}
Intent detection is a hard problem which is not completely solved yet. In this paper, we have applied reverse engineering techniques to the intent detection SNIPS dataset with promising results. It is worthy to highlight that highly interpretable low-dimensional representations of their state space were obtained, and we have shown that sentences seem to travel towards concrete regions of the space to generate the predictions. Moreover, unexpected structures of fixed points have been identified underlying the network. 

The work presented in this paper can be extended by following several lines, as exploring the role played by the fixed points and their basins of attraction, applying these ideas to other datasets to extend these results to a generic intent detection problem; or applying these techniques to other closely related problems as multi-intent and out-of-scope intent detection among many others. Finally, one particularly interesting line of work can focus on adapting this approach to Transformer architectures.

%%%%%%%%%%%%%%%%%%%%%%%%%%%%%%%%%%%%%%%%%%%%%%%%%%%%%%%%%%%%%%%%%%

%%%%%%%%%%%%%%%%%%%%%%%%%%%%%%%%%%%%%%%%%%%%%%%%%%%%%%%%%%%%%%%%%%%%%%%%

%%% Use this environment to include acknowledgements (optional).
%%% This will be omitted in doubleblind mode.

\begin{ack}
MAGN acknowledges the support by the European Union HORIZON-CL4-2021-HUMAN-01-01 under grant agreement 101070028 (REXASI-PRO) and by TED2021-129438B-I00 / AEI/10.13039/501100011033 / Unión Europea NextGenerationEU/PRTR.
%By using the \texttt{ack} environment to insert your (optional)  acknowledgements, you can ensure that the text is suppressed whenever  you use the \texttt{doubleblind} option. In the final version,  acknowledgements may be included on the extra page intended for references.
\end{ack}

%%%%%%%%%%%%%%%%%%%%%%%%%%%%%%%%%%%%%%%%%%%%%%%%%%%%%%%%%%%%%%%%%%%%%%%%

%%% Use this command to include your bibliography file.
\bibliography{m432}

\end{document}